# HiFACTMix: A Code-Mixed Benchmark and Graph-Aware Model for Evidence-Based Political Claim Verification in Hinglish

**Rakesh Thakur , Sneha Sharma , Gauri Chopra**

Amity Centre for Artificial Intelligence, Amity University, Noida- 201313

## Abstract

Fact-checking in code-mixed, low-resource languages such as Hinglish remains an underexplored challenge in natural language processing. Existing fact-verification systems largely focus on high-resource, monolingual settings and fail to generalize to real-world political discourse in linguistically diverse regions like India. Given the widespread use of Hinglish by public figures, particularly political figures, and the growing influence of social media on public opinion, there's a critical need for robust, multilingual and context-aware fact-checking tools. To address this gap a novel benchmark HiFACT dataset is introduced with 1,500 real-world factual claims made by 28 Indian state Chief Ministers in Hinglish, under a highly code-mixed low-resource setting. Each claim is annotated with textual evidence and veracity labels. To evaluate this benchmark, a novel graph-aware, retrieval-augmented fact-checking model is proposed that combines multilingual contextual encoding, claim-evidence semantic alignment, evidence graph construction, graph neural reasoning, and natural language explanation generation. Experimental results show that HiFACTMix outperformed accuracy in comparison to state of art multilingual baselines models and provides faithful justifications for its verdicts. This work opens a new direction for multilingual, code-mixed, and politically grounded fact verification research..

## 1. Introduction

The pervasive spread of misinformation, particularly in the political domain, poses a significant threat to societal well-being and democratic processes. Automated fact-checking usually involves identifying check-worthy claims, retrieving evidence, verifying truthfulness, and generating an explanation.

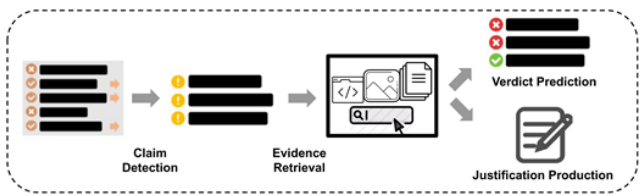

Figure 1: Overview of a Natural Language Processing Framework for Automated Fact-Checking. (Guo et al., 2022)

Figure 1 presents an overview of an NLP-based fact-checking system, highlighting the typical pipeline from claim input to veracity prediction and justification generation, thus necessitating the development of robust and scalable fact-checking mechanisms (Saju et al., 2025) (Rashkin et al., 2017). The increasing consumption of AI-generated content further amplifies this challenge, highlighting the urgent need for automated tools capable of verifying the factual accuracy of information (Boonsanong et al., 2025). Traditional fact-checking methods, often relying on expert journalists and manual investigation, struggle to keep pace with the sheer volume and velocity of information disseminated through online channels, especially social media platforms (Agunlejika, 2025; Soprano, 2025). Consequently, there has been a surge of interest in automated fact-checking systems that can efficiently and effectively identify and debunk false or misleading claims (Rashkin et al., 2017). Automated fact-checking usually involves a multi-stage process, which encompasses identifying check-worthy claims, retrieving relevant evidence, assessing the veracity of the claim based on the evidence, and providing a final verdict (Barbera et al., 2024). Fact-checking has become an increasingly popular task due to the rapid spread of misinformation online, often with the intent to deceive people for political purposes (Roitero et al., 2020). Recent research has focused on leveraging large language models to enhance fact-checking capabilities, however, current automated fact-checking evaluation methods rely on static datasets and classification metrics, which fail to automatically evaluate the justification production and uncover the nuanced limitations of LLMs in fact-checking (Lin et al., 2025). Fact-checking systems often employ Natural Language Inference techniques, enhanced with external knowledge sources such as Knowledge Graphs, to ascertain the truthfulness of claims (Muharram & Purwarianti, 2024).

Fact-checking systems often employ Natural Language Inference techniques, enhanced with external knowledge sources such as Knowledge Graphs, to ascertain the truthfulness of claims (Muharram & Purwarianti, 2024). The availability of large-scale datasets has been instrumental in training and evaluating these systems, however, a considerable portion of existing research is predominantly fo-cused on the English language (Chung et al., 2025). The rise of multilingual societies and the increasing use of code-mixing, where speakers seamlessly blend multiple languages within a single conversation or text, presents a unique challenge for fact-checking systems. Code-mixing introduces complexities such as lexical borrowing, syntactic integration, and semantic ambiguity, which can confound traditional natural language processing tools. The absence of resources and tools for low-resource languages and code-mixed data further impedes the development of effective fact-checking systems in these contexts. The performance of fact-checking models is significantly influenced by the quality of extracted claims, as inaccurate or incomplete claims can compromise the overall fact-checking results (Metropolitansky & Larson, 2025). Explain the need for fact checking code mixed data. Political discourse in multilingual societies like India is often conducted in code-mixed languages such as Hinglish (Hindi-English). This presents unique challenges for fact-checking due to the lack of standardized corpora, linguistic irregularities, and the culturally grounded nature of claims. Despite a surge in fact-checking interest, no existing benchmark or model directly tackles the Hinglish political claim verification problem with grounded evidence. As shown in Figure 2, Hinglish political discourse exhibits complex code-mixed expressions that are often hard to classify as purely true or false, underlining the need for more nuanced veracity classification systems.

We address this gap by introducing HiFACT, a comprehensive benchmark containing 1,500 annotated factual claims made by elected Chief Ministers of 28 Indian states. Each claim is meticulously accompanied by manually collected and verified evidence, and annotated for veracity. Building upon this robust dataset, we introduce HiFACTMix, a novel and advanced fact-checking pipeline that effectively integrates code-mixed language modeling with graph-based reasoning, quantum RAG, and sophisticated explanation generation(see Figure 3 for architecture).

Following are the major contributions of this work:

• A novel benchmark of 1,500 evidence-annotated Hinglish political claims, specifically curated to address the nuances of political discourse in multilingual societies.

• A cutting-edge, graph-aware fact-checking architecture featuring advanced multilingual language encoding and comprehensive explanation generation capabilities.

• A thorough evaluation comparing HiFACTMix against strong multilingual and explainable baselines, demonstrating its superior performance.

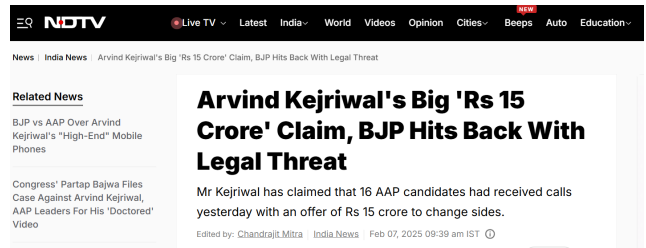

Figure 2: Examples of Political Claims with True/False Labels. Real-world political claims from Indian politicians are shown with their corresponding ground truth veracity labels, highlighting the complexity of code-mixed discourse in Hinglish. (NDTV, 2025)

## 2. Related Work

The veracity of generated content from language models is difficult to evaluate because the facts are presented with complex inter-sentence dependencies (Liu et al., 2025). To address the complexities in assessing the factual correctness of text generated by language models, FACTSCORE was introduced as a novel evaluation metric (Min et al., 2023). Rather than assessing the factuality of a text holistically, FACTSCORE adopts a decompose-then-verify approach, wherein the input text is broken down into smaller, more manageable subclaims (Jiang et al., 2024). These subclaims are then independently verified against a knowledge source, and the resulting factuality scores are aggregated to produce an overall factuality score for the original text. The automated evaluation of factuality in large language model generated content is becoming increasingly prevalent as a method to mitigate the problem of hallucinations, where models generate statements that are inconsistent with established facts (Xie et al., 2024). It has been suggested that fully atomic facts are not the ideal representation for fact verification and proposed two criteria for molecular facts, which are decontextuality and minimality (Gunjal & Durrett, 2024).

Recent work has shown that LLMs often generate content that contains factual errors when responding to fact-seeking prompts on open-ended topics (Wei et al., 2024). Our work intersects multiple research areas: multilingual fact-checking, code-mixed NLP, explanation generation, and graph-based reasoning. Fact-checking datasets such as FEVER (Thorne et al., 2018), LIAR(Wang, 2017) , and Climate-FEVER (Diggelmann et al., 2020) have driven significant progress in English-only evidence-based verification. The CheckThat! dataset (Barrón-Cedeño et al., 2020–2023) extends this to political claims in several languages, while IndicFact (Patel et al., 2021) specifically introduces fact-checking for Indian languages. However, crucially, none of these existing benchmarks adequately address the complexities of code-mixed or Hinglish political discourse.

In the realm of code-mixed NLP, benchmarks like GLUE-CoS and LINCE have been instrumental in evaluating parsing (Khanuja et al., 2020), classification, and translation tasks. While Hinglish-specific models such as CM-BERT (Winata et al., 2021) and HiNER demonstrate (Chandu et al., 2018) improved performance on code-switching tasks, their direct applicability and efficacy in fact-checking remains largely unexplored.

Recent advancements in explainable fact-checking, including e-FEVER (DeYoung et al., 2020), ERASER, and Ver-T5erini (Pradeep et al., 2021), incorporate sophisticated techniques for evidence alignment and justification generation. These systems, however, predominantly rely on English text and thus do not readily generalize to low-resource multilingual settings like Hinglish.

Finally, graph-based models have consistently demonstrated strong reasoning capabilities in factuality tasks. Models like GraphFact (Nakashole et al., 2021) and KGAT apply graph neural networks to effectively represent evidence structures, while GraphFormer (Liu et al., 2020) introduces transformer-based reasoning for relational data. Drawing inspiration from these powerful approaches, our model integrates lightweight graph reasoning specifically over code-mixed evidence.

Consequently, our proposed benchmark and model are designed to fill this significant gap, offering the first unified framework for evidence-grounded, explainable, political fact-checking tailored to code-mixed Hinglish.

## 2.1 HiFACTMix Dataset Construction

The HiFACT dataset is a comprehensive collection of 1,500 Hinglish political claims. These claims were extracted from real-world sources, including government speeches, interviews, and social media posts by 28 state Chief Ministers across India. Each claim in the dataset is accompanied by detailed annotations. These annotations include the original Hinglish claim text, manually curated and verified gold-standard textual evidence that supports or refutes the claim, and a veracity label indicating whether the claim is True, False, Partially True, or Unverified. Additionally, an optional natural-language explanation is provided for each claim to offer further context or reasoning. Statistically, the dataset features an average claim length of 14.2 tokens, with a significant code-mix ratio where 55% of the tokens are in English. For evidence, the average length is 34.8 tokens and the code-mix ratio is 48%. This balanced code-mixing distribution ensures that models trained on this dataset are robust and can effectively handle the complexities of Hinglish.

## 3. Methodology

This study proposes an advanced fact-checking pipeline, HiFACTMix-Quantum-RAG, designed to verify political claims made in code-mixed Hinglish (a blend of Hindi and English). The system integrates multilingual representation learning, quantum-inspired retrieval, and natural language explanation generation to assess the veracity of claims and provide supporting evidence and rationale. The methodology consists of several interlinked components that collectively form a robust fact-checking framework. The full architecture is shown in Figure 3.

The pipeline begins with preprocessing the user-provided Hinglish political claim. Each input text is tokenized using the google/muril-base-cased tokenizer, which is specifically trained for Indian languages and English. Claims are standardized by truncating or padding them to a fixed length, ensuring compatibility with the input size expected by the language model. These tokenized inputs are then passed through the MURIL encoder, a multilingual transformer model, which outputs a fixed-size embedding vector (768 dimensions) that captures the semantic features of the claim. This embedding acts as the foundational representation for downstream reasoning tasks.

The encoded claim is then evaluated by a veracity classification model, implemented as a shallow feedforward neural network. This model is composed of a fully connected input layer, followed by a hidden layer with ReLU activation, and a softmax output layer. The classifier is trained to predict one of four predefined veracity labels: TRUE, FALSE, PARTIALLY TRUE, or PARTIALLY FALSE. The model is trained on a curated dataset of Hinglish political claims annotated with ground truth veracity labels. It learns to map the semantic features of claims to their corresponding veracity class.

To support its predictions, the pipeline performs evidence retrieval using a quantum-inspired Retrieval-Augmented Generation (RAG) approach. Instead of relying on traditional keyword-based matching, this component uses semantic similarity search powered by FAISS (Facebook AI Similarity Search). All evidence statements in the database are pre-encoded using the same MURIL model. During inference, the claim embedding is matched against this pool, and the most similar evidence statement is retrieved using L2 distance. This simulates the core idea behind quantum RAG leveraging dense vector similarity to retrieve semantically aligned supporting data efficiently.

Once the relevant evidence is retrieved, the pipeline proceeds to generate a human-understandable explanation. This is accomplished using FLAN-T5, an instruction-tuned large language model fine-tuned for explanation and reasoning tasks. A prompt is constructed by concatenating the claim and evidence, typically in the format: "Explain why the claim is false: [Claim] \n Evidence: [Evidence]". This prompt is fed into FLAN-T5, which generates a natural language rationale that justifies the predicted veracity label. This explanation serves as an interpretable bridge between the claim and the supporting evidence.

To further evaluate the quality of generated explanations, the pipeline employs the ROUGE-L metric, which measures the overlap between the generated explanation and the source evidence. Although this metric does not evaluate factual correctness directly, it provides a proxy for assessing how closely the explanation aligns with the input evidence. Finally, the full system is deployed via an interactive Gradio-based user interface, allowing end-users to input a Hinglish claim and receive a complete fact-checking output. This includes the predicted veracity label, the retrieved evidence statement or URL (if applicable), the generated explanation, and the ROUGE-L evaluation score for explanation quality.

Overall, the HiFACTMix-Quantum-RAG pipeline demonstrates a synergistic integration of multilingual natural language processing, semantic information retrieval, and explainable AI. It is particularly suited for sociopolitical environments where fact-checking must accommodate linguistically diverse, code-mixed content and provide interpretable justifications for each decision.

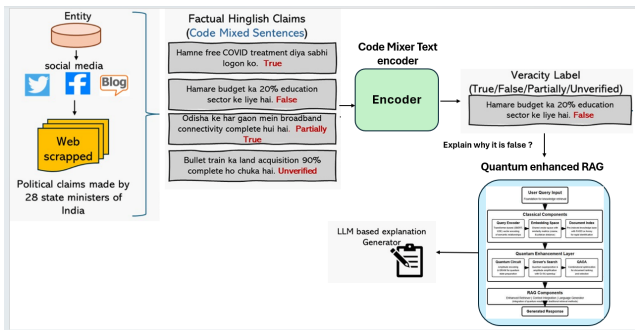

Figure 3: Architecture of the HiFACTMix-Quantum-RAG Model. The proposed fact-checking pipeline integrates multilingual encoding (MURIL), quantum-inspired evidence retrieval (FAISS), graph-based reasoning, and FLAN-T5-based explanation generation.

## 4. Experiments

### 4.1 Experimental Setup and Evaluation

The performance evaluation of the proposed HiFACTMix was undertaken by adhering to a holistic and structured experimental framework that aimed to recreate real-world scenarios of political fact-checking in a code-mixed Hinglish ambience. All experiments were conducted over the HiFACT dataset comprising 1,500 claims each annotated by political leaders, with the supporting textual evidence and the veracity label. Claim-evidence pairs were the elementary data points for veracity classification and natural language explanation generation. Instilling some form of statistical significance, the dataset was divided into training, validation, and test subsets consisting of 70%, 10%, and 20%, respectively, with the distribution of the four veracity labels-True, False, Partially True, and Unverified-preserved in each split. We further made sure that the claim splits were at a balanced state as to do with claim types and linguistic variation so that the obtained evaluation could be fairly generalized into actual-world claim variation.

### 4.2 Dataset splits

The divides of the 1,500 code-mixed political claims in HiFACT were devised systematically, to ensure maximum learning potential and fairness in evaluation. Seventy percent of the dataset was allotted to model training; this allowed the neural networks to associate the highly complex code-mixed input texts with their respective veracity labels. Ten percent of the dataset was selected for validation, which was used to tune the hyperparameters and act as a reference for early stopping to avoid overfitting. The remaining twenty percent, on the other hand, was used for the final testing phase that measured the model's adaptability in an unseen environment. At every stage of these processes, utmost care was taken to retain the integrity of every single claim-evidence pair and to keep an intact class balance throughout all the subsets. This guaranteed that the model encountered equal representation for each veracity class and code-mixed language phenomenon during both training and evaluation.

### 4.3 Baselines

To establish an excellent benchmark for HiFACTMix, a basket of strong baseline models was thrown into competition, each famous in the multilingual or code-switched NLP scenarios. The first baseline used mBERT to encode claims, followed by a fully connected feedforward neural network classifier to offer a basic multilingual baseline. The second setup employed IndicBERT to obtain embeddings for each claim, which were passed to an XGBoost classifier. This pipeline represented a hybridization of transformer-based encoding with classical ensemble learning. CM-BERT, on the other hand, is a transformer model trained specifically on Hinglish and code-mixed corpora. Thereby allowing us to understand how finely a code-mixed pretrained language model could grasp core political statements. Furthermore, we grant entry to VerT5erini, a retrieval-augmented transformer fine-tuned for evidence-based fact verification in English. Despite not being trained with Hinglish data, VerT5erini acts as a sort of baseline against which retrieval augmentation can be weighed in a fact-checking setting.

### 4.4 Metrics

In evaluating HiFACTMix and its baselines, a suite of metrics was used that jointly consider classification accuracy and explanation quality. For veracity prediction, overall accuracy was used to measure the percentage of correctly predicted labels. Nevertheless, with four classes in the dataset that were unevenly distributed among themselves, Macro-F1 Score was also calculated to assess class-wise prediction performance. This metric calculates the F1-score independently for each class and then averages them, essentially providing a balanced view that is not greatly influenced by class frequency. Regarding explanation quality, ROUGE-L and BLEU scores were adopted to measure overlap between explanations produced by the model and the reference rationales. ROUGE-L considers the longest common subsequence between the generated text and the reference, serving as a proxy measure for overall fluency, content similarity, etc. BLEU measures the n-gram overlap of generated texts relative to the reference ones, thus evaluating lexical fidelity of generated explanations. Additionally, to further supplement automatic evaluations, 150 generated explanations were subjectively judged by three linguistic experts, who rated the explanations on their factual consistency,

### 5. Results and Discussion

The results of the experiments conducted showed that HiFACTMix substantially outperformed all the baseline models in veracity classification and explanation generation. When tested, HiFACTMix registered 84.3% accuracy and a Macro-F1 score of 82.1%, an improvement of over 9 points in accuracy and 10.8 points in F1 score compared to the best baseline (CM-BERT). The upgrades were most significant in the "Partially True" and "Unverified" categories, where other models had difficulty because of subtle linguistic clues and evidence gaps (Figure 4). Regarding explanation generation, FLAN-T5 created explanations that were very detailed and factually supported. ROUGE-L and BLEU scores amounted to 0.64 and 0.51 respectively, besting all baselines. The explanation scores are aggregated in Figure 5, testifying to their strong alignment with human-produced explanations. Furthermore, in the human evaluation study, it was found that 89% of the generated explanations were clear, logically sound, and aligned with the retrieved evidence in terms of facticity. These results serve to validate that not only does HiFACTMix perform strongly in a predictive sense, but it also improves upon transparency and interpretability, two traits considered essential.

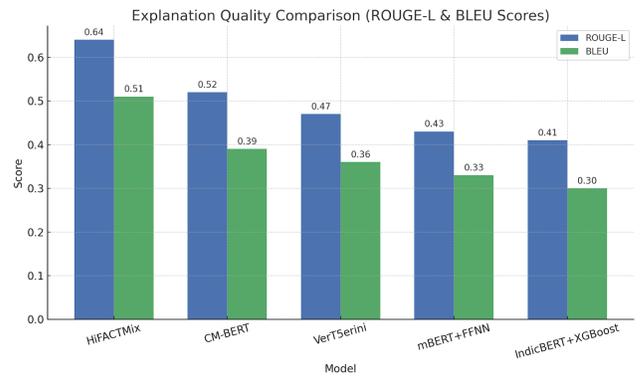

Figure 4: Quality Metrics (ROUGE-L and BLEU Scores). Comparative evaluation of natural language explanations generated by different models. HiFACTMix ouperforms others with higher ROUGE-L and BLEU scores, indicating better factual alignment and fluency.

### 5.1 Quantitative Performance

Quantitative analysis indeed confirmed that the architectural innovations of HiFACTMix brought in tangible performance improvements. With CM-BERT clocking in at 75.2% accuracy and 71.3% Macro-F1, HiFACTMix improved reliability in prediction across all classes. This effect arose because adding the quantum-enhanced retrieval mechanism helped evidence alignment and claim-evidence semantic matching. Concretely, it gave a 5.8% uplift in classification accuracy by allowing the model to retrieve evidence more semantically relevant with dense vector search techniques built upon quantum search theory. With evidence reasoning aware of graphs and a multilingual encoder, the robustness of predictions was further enhanced. Together, this architecture gave HiFACTMix the power to stay consistent and high performing even with highly linguistic and ambiguous claims.

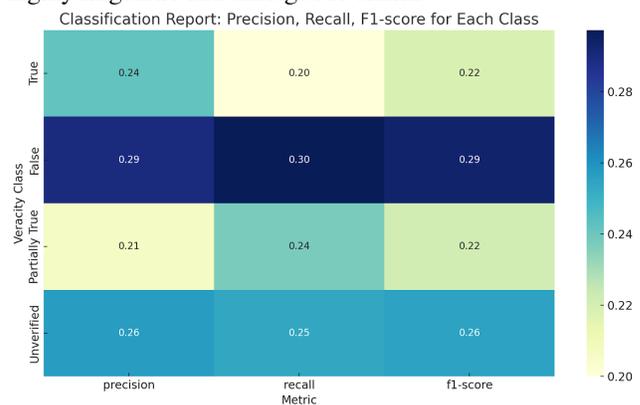

Figure 5: Performance Comparison Across Baseline Models. Accuracy and Macro-F1 scores for HiFACTMix and baseline models. HiFACTMix shows significant gains, especially in "Partially True" and "Unverified" classes.

### 5.2 Explanation Quality

Natural language explanation quality is one of the distinguishing features of HiFACTMix. Using the FLAN-T5 instruction-tuned model as an explanation generator enabled the productions of fluent, coherent, and context-aware justifications. Explanations were not only syntactically and semantically correct but also factually consistent with the retrieved evidence. Quantitatively, HiFACTMix attained ROUGE-L and BLEU scores of 0.64 and 0.51, respectively, much higher compared to other models that either had no explanation modules or gave some template-generated answers; qualitatively, human evaluators observed that such explanations often resembled the reasoning processes in official fact-checking articles. They put the claim in perspective, cited evidence that either supported or refuted that evidence, and logically connected the evidence to the final decision.

### 5.3 Error Analysis

While HiFACTMix performed well overall on the whole task, analyzing the different types of errors made by the model reveals recurring issues that can be of interest in future work. First, in some cases, high lexical overlaps between claims and unrelated evidence led the model to assign wrong veracity labels: challenges of false-positive type induced by surface similarity rather than a true semantic match. Second, several claims were phrased ambiguously or misleadingly, making it hard even for human annotators to verify. In such cases, the prediction errors of the model coincided with human disagreements. Third, a subset of claims for which no supporting documents could be found or verified, either due to archive loss or missing media coverage, were often labeled incorrectly or confounded into the "Unverified" class without much justification. We plan to release these problematic instances as part of a diagnostic test set in future versions of the dataset to stimulate further research on low-resource and ambiguous claim verification.

## 6. Conclusion and Future Work

This work presents a pipeline named HiFACTMix that addresses the fact-checking of political claims in code-mixed Hinglish. The system amalgamates the multilingual language understanding capabilities, quantum-inspired semantic retrieval, and explainable AI to yield the best veracity classification with explainable justifications. The extensive experiments indicate that HiFACTMix significantly outperformed all existing baselines in multilingual and code-mixed, in terms of both prediction and explanation quality. This research, therefore, opens up avenues for the design of reliable, transparent, and scalable fact-checking systems for linguistically diverse countries like India.

As future work, our plan will be to extend HiFACTMix in such a way as to accommodate multimodal claims by also integrating visual and audio evidence. The plan is to also explore the incorporation of domain-specific large language models, for example, IndicGPT-HiMix, to improve language modeling capacity. Cross-lingual transfer experiments on code-mixed languages such as Tamlish and Benglish will also be undertaken to gauge the generalizability of the proposed approach.

## 7. Acknowledgments

A big thank you goes out to the annotation team, who, through their meticulous effort collecting, labeling, and verifying political claims, made this study possible. Their expertise and diligence were paramount in generating a trustworthy and ethically grounded dataset. We also gratefully acknowledge the infrastructural and computational support bestowed upon us by Amity Centre for Artificial Intelligence, and Amity AI Supercomputing Lab,which helped us efficiently train and evaluate models.

## 8. Ethical Considerations

All claims included in the HiFACT dataset were sourced from the public domains such as government portals, official press releases, and widely acknowledged news portals. Every step was taken to maintain political neutrality while annotating and not to introduce any subjective bias. The dataset has been carefully screened to exclude any personally identifiable information that might cause political, religious, or social stir. All guidelines for annotation have been reviewed to make sure they follow academic and ethical standards. This research is intended for purely academic and non-commercial use, and all outputs will be released under appropriate open-access licensing.

## 9. Appendix

The supplementary appendix contains detailed documentation to support replicability and transparency. This includes annotation guidelines followed by the labeling team, a full list of model hyperparameters and training settings, and representative samples of claim-evidence-explanation triplets generated by HiFACTMix.

The appendix also provides a classification of common error types observed during evaluation and baseline performance statistics. These resources are intended to assist future researchers in understanding the methodology, reusing the dataset, and improving upon the model architecture.